\documentclass[11pt]{article}

\usepackage[utf8]{inputenc}
\usepackage[T1]{fontenc}
\usepackage{newtxtext,newtxmath}
\usepackage{amsmath}

\usepackage{amssymb,amsfonts}
\usepackage{graphicx}
\usepackage{booktabs}
\usepackage{multirow}
\usepackage{adjustbox}
\usepackage{microtype}
\usepackage{xcolor}
\usepackage{float}
\usepackage{enumitem}
\usepackage{geometry}
\usepackage{tikz}
\usepackage{pgfplots}
\pgfplotsset{compat=1.18}
\usepgfplotslibrary{groupplots}
\usetikzlibrary{arrows.meta,positioning,calc,fit,backgrounds,shapes.geometric}
\usepackage[authoryear,round]{natbib}
\usepackage{xurl}
\usepackage{hyperref}
\usepackage{cleveref}
\usepackage{CJK}
\newcommand{\GitHubReleaseURL}{\url{https://github.com/SymbolicLight-AGI/SymbolicLight-V1}}
\newcommand{\HFReleaseURL}{\url{https://huggingface.co/SymbolicLight-AGI/SymbolicLight-V1}}

\hypersetup{
  colorlinks=true,
  linkcolor=blue!55!black,
  citecolor=blue!55!black,
  urlcolor=blue!55!black,
  pdftitle={SymbolicLight V1: Spike-Gated Dual-Path Language Modeling at High Encoder Spike Sparsity},
  pdfauthor={Ting Liu},
  pdfkeywords={spiking neural networks, language models, activation sparsity, neuromorphic computing, surrogate gradient, dual-path architecture, scale-up evidence}
}

\definecolor{spikered}{HTML}{E74C3C}
\definecolor{spikeblue}{HTML}{2980B9}
\definecolor{spikegreen}{HTML}{27AE60}
\definecolor{spikepurple}{HTML}{8E44AD}
\definecolor{spikeorange}{HTML}{E67E22}
\definecolor{spikegray}{HTML}{4A5568}
\definecolor{decaycolor}{HTML}{3498DB}
\definecolor{attncolor}{HTML}{E74C3C}
\definecolor{lightblue}{HTML}{EBF4FF}
\definecolor{lightorange}{HTML}{FFFAF0}
\definecolor{lightgreen}{HTML}{F0FFF4}
\definecolor{lightpurple}{HTML}{FAF5FF}
\definecolor{blockbg}{HTML}{F7FAFC}

\title{%
  \textbf{SymbolicLight V1: Spike-Gated Dual-Path \\
  Language Modeling at High Encoder Spike Sparsity}
}

\author{
  Ting Liu \\
  SymbolicLight Research \\
  Foshan, Guangdong, China \\
  \texttt{research@symboliclight.com}
}

\date{September 2026}

\begin{document}
\maketitle

\begin{itemize}[leftmargin=1.4em,itemsep=1pt,topsep=2pt,parsep=0pt]
  \item Public code repository: \GitHubReleaseURL.
  \item Model artifacts: \HFReleaseURL.
\end{itemize}

\begin{abstract}
Natively trained spiking language models must preserve information across time while operating through sparse binary activations, a combination that has produced a persistent quality gap relative to dense Transformers.
We present \textbf{SymbolicLight V1}, a spike-gated dual-path language model that couples binary Leaky Integrate-and-Fire (LIF) dynamics with a continuous residual stream.
Its Dual-Path SparseTCAM mixer combines a first-order exponential-decay state with windowed local attention on the continuous residual stream, followed by a context-conditioned decoding head.

We train four 194M-parameter models from scratch on a 3B-token, 10-domain Chinese--English corpus.
On a fixed token-weighted evaluation set the runs reach perplexity (PPL) 8.88--8.93 (mean 8.904, sample standard deviation 0.019). Separation of this set from the training streams has not been verified. Training-time encoder probes have a mean zero-spike fraction of 89.96\%; this is not a whole-model sparsity measure.
Code tokens are 43.7\% of that set; the unweighted mean of the ten domain PPLs is 29.38.
Under the same corpus, tokenizer, token budget, and hardware, the token-weighted mean is 7.7\% above GPT-2 201M (PPL 8.27).
Across five zero-shot benchmarks the two 200M-scale models show no clear accuracy separation.
Under sampling with temperature 0.7 and top-$k$ 50, SymbolicLight produces lower 4-gram repetition; an entropy-modulated rule reverses that ranking.
On an RTX 2080 Ti, measured generation throughput is 22.8 versus 91.5 token/s; post-generation power readings give rough energy estimates of 2,848 versus 905\,mJ/token, without power integration over generation.
\end{abstract}


\section{Introduction}
\label{sec:intro}

\subsection{The Quality Gap of Native Spiking Language Models}

Natively trained spiking neural networks (SNNs), trained from random initialization under surrogate gradients~\citep{neftci2019surrogate}, have historically struggled to match dense Transformer~\citep{vaswani2017attention} quality on autoregressive language modeling.
The gap is not only that activations are binary: the model must carry information across time through sparse events, and the failure becomes clearer once pre-training leaves narrow synthetic text for multi-domain corpora with heterogeneous token distributions.
Conversion-based and quantization-based spiking LMs build on pretrained dense networks~\citep{xu2025nsllm,xing2024spikellm}; native models such as SpikeGPT~\citep{zhu2023spikgpt} combine RWKV-style recurrence, continuous internal states, and spiking activations.
Sub-quadratic dense architectures such as Mamba~\citep{gu2023mamba} and RWKV~\citep{peng2023rwkv} reduce mixing cost without binarizing activations.
Biological circuits combine sparse spikes with continuous internal dynamics~\citep{maass1997networks,roy2019towards}, which motivates a hybrid design but does not by itself close the quality gap.

\subsection{Key Insight: Spike-Gated Dual-Path Computation}

SymbolicLight V1 adopts a hybrid view of spiking computation.
Binary LIF spikes supply sparse-valued inputs to selected projections, while a continuous residual stream carries representational content. The current implementation uses dense kernels; zero spike values do not themselves skip arithmetic or memory accesses.
This design is consistent with the fact that biological circuits combine discrete action potentials with continuous dendritic integration~\citep{london2005dendritic,gerstner2002spiking}.

The resulting architecture uses spikes as event gates rather than forcing every computation to be one-bit.
Dual-Path SparseTCAM combines a first-order exponential-decay state, whose learned half-life is about 8--13 tokens, with windowed local attention on the continuous residual stream, retaining sparse encoder outputs while using continuous-valued sequence mixing.

\subsection{Contributions}

We report three contributions:

\begin{enumerate}[leftmargin=*,itemsep=2pt]
  \item \textbf{A SymbolicLight V1 spike-gated dual-path language model.}
  The architecture combines LIF spike dynamics, Dual-Path SparseTCAM, a context-conditioned MLP head, and a model-specific bilingual tokenizer.
  Binary spikes feed the decay-path projection and the FFN down-projection; the FFN up-projection and local softmax attention consume continuous inputs.

  \item \textbf{Native pre-training with sparse encoder outputs.}
  Four 194M-parameter runs trained from scratch on a 3B-token, 10-domain bilingual mixture reach token-weighted evaluation PPL 8.88--8.93 (mean 8.904, sample standard deviation 0.019) with mean sampled encoder zero-spike fraction 89.96\%.
  Code tokens are 43.7\% of the evaluation set; the unweighted mean of the ten domain PPLs is 29.38.

  \item \textbf{Scale-up and public inference artifacts.}
  A 0.8B-parameter run trained for 48.8B tokens shows that the same stack remains trainable at sub-billion scale and yields a loadable checkpoint.
  We release checkpoints, tokenizer assets, parameter-compatible model definitions, and minimal inference programs for loading and generation.
\end{enumerate}

\section{Related Work}
\label{sec:related}

\paragraph{Conversion-Based SNN Language Models.}
NSLLM~\citep{xu2025nsllm} combines spiking representations with quantization, sparsification, and algorithm--hardware co-design, including FPGA evaluation. SpikeLLM~\citep{xing2024spikellm} applies saliency-based spiking to pretrained large language models. Their use of pretrained dense models differs from the random-initialization training studied here; NSLLM also directly investigates hardware efficiency.

\paragraph{Natively-Trained SNN Language Models.}
SpikeGPT~\citep{zhu2023spikgpt} combines RWKV-style recurrence with continuous internal states and spiking activations, including a 216M-parameter model.
An earlier SymbolicLight prototype~\citep{liu2026symboliclight} introduced spike-gated associative lookup and LIF spike encoding; a follow-up preprint~\citep{liu2026scaling} scaled native pre-training to a multi-domain mixture.
These preprints report high activation sparsity together with quality gaps against dense baselines; their comparisons do not isolate binarization as the cause.
SymbolicLight V1, as presented here, unifies these design lessons in a spike-gated dual-path architecture that incorporates continuous local attention and a context MLP head while preserving spike-conditioned sparsity on the decay and FFN paths.
SpikeGPT evaluates on Enwik8, WikiText-2, and WikiText-103, with character-level and BPE tokenization in different experiments and OpenWebText2 pre-training for the 216M model. These corpora, tokenizers, and the fine-tuned WikiText evaluation protocol differ from ours, so their PPL values are not directly comparable.
Architecturally, SymbolicLight V1 differs from SpikeGPT in three ways: (i) windowed local attention on continuous Q/K/V replaces the RWKV-style recurrent mixer, (ii) a learnable gate fuses that attention path with a first-order decay state, and (iii) a context MLP on the residual stream supplements the vocabulary projection.

\paragraph{Linear Attention and Linear-Time Sequence Models.}
Linear attention variants such as RetNet~\citep{sun2023retentive} and GLA~\citep{yang2024gated} replace softmax attention with recurrent state updates; at fixed model width, recurrence over $S$ tokens takes $O(S)$ work and retains state independent of $S$.
Linear-time sequence models such as Mamba~\citep{gu2023mamba} and RWKV~\citep{peng2023rwkv} achieve competitive quality with linear complexity.
Our exponential-decay aggregation path shares mathematical structure with these approaches (a first-order linear recurrence), but differs in using \emph{binary spike gates} to control information flow, producing sparse-valued inputs to the decay projection. The measured 89.96\% zero-spike fraction refers specifically to sampled encoder outputs.

\paragraph{Hybrid Sparse Attention.}
Longformer~\citep{beltagy2020longformer} and BigBird~\citep{zaheer2020bigbird} combine sliding-window local attention with global tokens.
Our Dual-Path SparseTCAM combines sliding-window local attention on the continuous residual stream with per-element sparse representations on the decay path.
The position-level attention mask almost never excludes a token (\Cref{sec:sparse_attn}); the local window restricts which weights may be nonzero, but the released implementation still computes a dense attention matrix.

\paragraph{Extreme Quantization.}
BitNet b1.58~\citep{ma2024era} restricts weights to ternary states $\{-1, 0, 1\}$ while using 8-bit activations, rather than binarizing the residual stream.
Our hybrid model instead uses binary $\{0,1\}$ activations at selected interfaces and a recurrent membrane state in the encoder; its residual stream remains continuous. This differs from BitNet\textquotesingle s joint weight and activation quantization.
Notably, our parameter budget analysis (\Cref{tab:param_budget}) shows that the additional model-specific components (decay path, context MLP, fusion gate) consume only ${\sim}$12\% of total parameters; the architectural novelty lies in how these components interact with standard building blocks, not in parameter count.

\paragraph{Energy-Efficient Transformers.}
Knowledge distillation transfers predictive behavior to a smaller model~\citep{hinton2015distilling}, while the lottery-ticket study uses iterative pruning and retraining to identify sparse trainable subnetworks~\citep{frankle2019lottery}.
SNNs provide a complementary, architecture-level approach: sparsity emerges naturally from spiking dynamics during training rather than being imposed during compression.
Encoder probes average 89.96\% zero spikes without explicit sparsity regularization; individual logged probes can be lower (\Cref{sec:training_dynamics}).

\paragraph{Biological Neural Computation.}
Cortical neurons communicate through spike-gated continuous dynamics: binary action potentials gate synaptic transmission, while graded potentials in dendrites perform analog integration~\citep{london2005dendritic,gerstner2002spiking}.
Our spike-gated dual-path architecture is motivated by an analogy to this biological organization: binary LIF spikes gate computation flow, while a continuous residual stream preserves representational content.

\paragraph{Auxiliary Losses and Deep Supervision.}
CALM~\citep{schuster2022confident} and DeeBERT~\citep{xin2020deebert} add classifiers at intermediate layers for adaptive inference.
Our AuxCE applies exponentially-decayed auxiliary cross-entropy at every block during \emph{training} as deep supervision.

\section{Model Architecture}
\label{sec:arch}

\subsection{Architectural Positioning}
\label{sec:positioning}

We characterize SymbolicLight V1 as a \textbf{spike-gated dual-path} architecture (\Cref{fig:arch}).
Binary spikes feed the decay-path projection and the FFN down-projection; the FFN up-projection and local softmax attention consume the continuous residual stream $\mathbf{c}$.
Every ten optimizer steps, the training logger probes the encoder on the first 32 tokens of one sample. Across four runs, the 2,288 recorded probes have a mean zero-spike fraction of 89.96\%. The denominator is the $32\times768$ encoder-output elements in each probe, rather than all model activations or all tokens in the batch. These probes do not establish FFN intermediate sparsity or the number of executed operations.
The reported token-level all-zero rate is $<10^{-6}$, so the position-level mask is effectively inactive in the measured setting. It does not provide token-skipping evidence.

The architecture retains continuous-valued components for training stability and representational fidelity: LayerNorm, a GELU-activated context MLP, and softmax in the local attention computation.
The continuous residual path $\mathbf{c} \in \mathbb{R}^{B \times S \times D}$ serves as a gradient conduit.

Earlier SymbolicLight prototypes~\citep{liu2026symboliclight,liu2026scaling} reported high activation sparsity alongside quality gaps against dense baselines, without isolating the contribution of activation binarization.
This unified V1 design tests whether a continuous residual stream and windowed local attention reduce that quality gap while keeping spike-gated sparsity on the decay and FFN paths.
The resulting architecture is a hybrid spike-gated dual-path model, not a pure SNN.
\Cref{tab:bio_analogy} lists design analogies between architectural components and biological processes; a more detailed discussion of the biological grounding appears in \Cref{sec:discussion}.

\begin{table}[htbp]
\centering
\caption{Design analogies of SymbolicLight components; these are not validations of biological mechanisms or hardware efficiency.}
\label{tab:bio_analogy}
\small
\begin{adjustbox}{max width=\linewidth}
\begin{tabular}{@{}p{0.25\linewidth}p{0.60\linewidth}@{}}
\toprule
\textbf{Architecture Component} & \textbf{Biological Analogy} \\
\midrule
LIF spike output $s \in \{0,1\}$ & Axonal action potential (binary, all-or-none) \\
Continuous residual path $\mathbf{c}$ & Dendritic integration (graded potentials) \\
Exponential decay aggregation & Passive leaky integration (learned, activity-independent decay per head) \\
\bottomrule
\end{tabular}
\end{adjustbox}
\end{table}

The overall data flow is:
\begin{equation}
  \mathbf{x} \xrightarrow{\text{SpikeEncoder}} (\mathbf{s}, \mathbf{c}) \xrightarrow{\text{Blocks} \times L} (\mathbf{s}', \mathbf{c}') \xrightarrow{\text{Context MLP head}} \mathbf{y},
\end{equation}
where $\mathbf{s} \in \{0,1\}^{B \times S \times D}$ are binary spikes and $\mathbf{c} \in \mathbb{R}^{B \times S \times D}$ are continuous representations; $B$ is batch size, $S$ sequence length, $D$ hidden width, and $L$ the number of blocks. In the equations below, vectors are columns and batch and block indices are omitted unless needed.

\begin{figure}[htbp]
\centering
\begin{tikzpicture}[>=Stealth, font=\small,
 box/.style={draw=spikegray,rounded corners,align=center,minimum height=0.7cm,fill=white},
 ar/.style={->,thick}, signal/.style={font=\scriptsize,fill=white,inner sep=1pt}]
\node[box,fill=lightblue,minimum width=10cm] (enc) at (0,0) {Token embedding + LayerNorm + stateful LIF encoder};
\node[box,fill=lightgreen,minimum width=3.7cm] (dec) at (-2.4,-1.6) {Binary-input projection\\+ exponential-decay state};
\node[box,fill=lightorange,minimum width=3.7cm] (att) at (2.4,-1.6) {Continuous Q/K/V\\+ masked local softmax};
\node[box,minimum width=7.8cm] (mix) at (0,-3) {$g\,\mathrm{Attn}+(1-g)\,\mathrm{Decay}$; output projection};
\node[box,minimum width=7.8cm] (res) at (0,-4.1) {Add continuous residual + LayerNorm: $\mathbf{u}$};
\node[box,fill=lightpurple,minimum width=7.8cm] (ffn) at (0,-5.25) {Continuous up-projection $\to$ threshold $\to$ binary $\mathbf{b}$\\$\to$ down-projection; add $\mathbf{u}$ + LayerNorm};
\node[box,minimum width=7.8cm] (next) at (0,-6.5) {Next block receives $\mathbf{c}'$ and $\mathbf{s}'=\Theta(\mathbf{c}'-\theta)$};
\node[box,fill=lightblue,minimum width=7.8cm] (head) at (0,-7.7) {After 12 blocks: continuous $\mathbf{c}'$ $\to$ context MLP head};
\draw[ar] (enc.south -| dec) -- node[signal]{$\mathbf{s}$} (dec);
\draw[ar] (enc.south -| att) -- node[signal]{$\mathbf{c}$} (att);
\draw[ar] (dec) -- (mix.north -| dec);
\draw[ar] (att) -- (mix.north -| att);
\draw[ar] (mix)--(res);\draw[ar] (res)--(ffn);\draw[ar] (ffn)--(next);\draw[ar] (next)--(head);
\draw[ar,spikeblue] (enc.east) -- (5.5,0) |- node[signal,pos=0.75]{$\mathbf{c}$ residual} (res.east);
\draw[dashed,spikegray,rounded corners] (-4.3,-0.9) rectangle (4.3,-6.95);
\node[signal] at (-2.4,-6.95) {One repeated block};
\end{tikzpicture}
\caption{Signal flow of the 194M hybrid model. Binary spikes feed only selected projections. The continuous stream supplies attention, the FFN up-projection, residual additions, and the decoding head, and passes between blocks. FFN thresholds are stateless; the encoder LIF retains membrane state. Dense kernels still execute the major projections.}
\label{fig:arch}
\end{figure}

\subsection{SpikeEncoder with Chunked Sequential Processing}
\label{sec:encoder}

Token embeddings are layer-normalized to give the continuous input $\mathbf{x}_t=\mathbf{c}^{(0)}_t$. The LIF encoder separates the pre-reset potential $\mathbf{u}_t$ from the carried post-reset state $\mathbf{v}_t$:
\begin{align}
 \mathbf{u}_t &= \operatorname{clip}(\beta\mathbf{v}_{t-1}+\mathbf{x}_t,-3,3),\\
 \mathbf{s}_t &= \Theta(\mathbf{u}_t-\theta),\qquad
 \mathbf{v}_t=\mathbf{u}_t\odot(1-\mathbf{s}_t).
\end{align}
Here $\beta=0.95$ and $\theta=1$ are fixed, $\Theta(a)=\mathbf{1}[a\geq0]$ acts elementwise, and $\odot$ denotes elementwise multiplication. A fresh independent sequence starts with $\mathbf{v}_{-1}=\mathbf{0}$. Backpropagation uses the surrogate in \Cref{sec:atan}. The encoder returns both the binary spikes and continuous normalized embeddings. The fixed encoder leak is distinct from the learned per-head decay factors below.

\paragraph{Chunked Sequential Processing.}
The length-$S$ sequence is divided into chunks of size $C{=}64$.
Within each chunk, the LIF update is computed sequentially; across chunks, the terminal membrane potential is forwarded as the initial state of the next.
Chunking structures state handoff and bounds the size of intermediate spike lists, but the implementation still performs one sequential Python-level LIF update for each of the $S$ time steps.

\paragraph{RoPE Decoupling.}
Position encoding is removed from the SpikeEncoder.
Position information is injected downstream via Rotary Position Encoding (RoPE)~\citep{su2021roformer} applied to Q/K projections inside SparseLocalAttention (\Cref{sec:dual_path}), keeping the residual stream free of accumulated rotational artifacts.

\subsection{Dual-Path SparseTCAM}
\label{sec:dual_path}

We call the sequence mixer SparseTCAM, a name inspired by ternary content-addressable memory (TCAM)~\citep{pagiamtzis2006content}; it is implemented with learned projections, not a physical associative-memory lookup. It combines two pathways: a first-order decay state driven by spike-masked projections, and windowed local attention on the continuous residual stream.

\subsubsection{Path 1: Exponential Decay State (First-Order Recurrence)}
Each of $H{=}12$ heads maintains a hidden state $\mathbf{h}_h$ updated via:
\begin{equation}
  \mathbf{h}_h^{(t)} = \alpha_h \cdot \mathbf{h}_h^{(t-1)} + (1 - \alpha_h) \cdot \mathbf{z}_h^{(t)},
\end{equation}
Here $\alpha_h=\sigma(\gamma_h)$, with $\sigma$ the logistic sigmoid and $\gamma_h$ learned per head. The input is $\mathbf{z}^{(t)}_h=(W_{\rm dec}\mathbf{s}_t)_h\in\mathbb{R}^{d_k}$, where the subscript selects a head of width $d_k=D/H$. A fresh sequence starts with $\mathbf{h}^{(-1)}_h=\mathbf{0}$; concatenating the head states gives $\operatorname{Decay}_t\in\mathbb{R}^D$.
Excluding the input projection, the state update admits $O(SD)$ work for $S$ tokens and $O(D)$ streaming state. Incremental decoding applies this update directly. The released full-sequence path instead uses a length-$S$ grouped convolution; a linear-work scan is not implemented there.
Learned $\alpha$ values correspond to half-lives of about 8--13 tokens (\Cref{app:perhead_decay}); after 256 steps the residual weight is negligible, so the decay path does not implement long-range memory at the trained operating point.

\subsubsection{Path 2: Windowed Local Attention}
\label{sec:sparse_attn}

For head $h$, the continuous stream supplies
\begin{align}
 \mathbf{q}_{h,t}&=R_tW_{Q,h}\mathbf{c}_t,\qquad
 \mathbf{k}_{h,t}=R_tW_{K,h}\mathbf{c}_t,\label{eq:q_proj}\\
 \mathbf{v}_{h,t}&=W_{V,h}\mathbf{c}_t,\label{eq:v_proj}
\end{align}
where $W_{Q,h},W_{K,h},W_{V,h}\in\mathbb{R}^{d_k\times D}$ and $R_t$ is the head-wise RoPE rotation. Positions are zero-indexed. Define the active-position indicator and allowed keys by
\begin{align}
 m_t&=\mathbf{1}[\|\mathbf{s}_t\|_1>0],\\
 \mathcal{K}_t&=\{j:0\leq j\leq t,\ ((t-j\leq w)\lor(j<N_g)),\ m_j=1\}.
\end{align}
Causality applies to both the local window and the $N_g=4$ initial global anchors. The window parameter is the maximum lag: $w=256$ permits the current token and up to 256 previous tokens before union with the anchors. The 0.8B model uses $w=512$, covering its full training context. The attention output is
\begin{equation}
 \operatorname{Attn}_{h,t}=
 \begin{cases}
 m_t\displaystyle\frac{\sum_{j\in\mathcal{K}_t}\exp(\mathbf{q}_{h,t}^{\top}\mathbf{k}_{h,j}/\sqrt{d_k})\mathbf{v}_{h,j}}
 {\sum_{j\in\mathcal{K}_t}\exp(\mathbf{q}_{h,t}^{\top}\mathbf{k}_{h,j}/\sqrt{d_k})},&\mathcal{K}_t\ne\varnothing,\\[6pt]
 \mathbf{0},&\mathcal{K}_t=\varnothing.
 \end{cases}\label{eq:attention}
\end{equation}
Concatenating heads gives $\operatorname{Attn}_t\in\mathbb{R}^{D}$. Thus inactive keys are excluded, inactive queries have zero output, and an empty key set produces zero rather than an undefined softmax.

The reported token-level all-zero rate is $<10^{-6}$, so the position mask rarely excludes keys in the reported setting. An implementation that visits only the permitted keys could use $O(S(w+N_g)d_k)$ attention work per head. The released code instead forms all $S\times S$ scores, masks them, and performs a dense weighted sum: full-sequence attention therefore requires $O(S^2d_k)$ arithmetic per head and $O(S^2)$ score storage. Cached decoding appends all historical K/V entries without window eviction; its per-token attention work and cache storage grow with the accumulated sequence length.
Spike-driven sparsity enters through the decay-path projection on $\mathbf{s}$ and through the stateless threshold at block boundaries, not through this mask.
The 89.96\% zero-spike fraction is measured by encoder probes (\Cref{sec:training_dynamics}).

\subsubsection{Gated Fusion}
The two paths are fused via a learnable scalar gate $g = \sigma(w_g)$, defined separately in each block and initialized at $w_g{=}0$ for equal mixing coefficients:
\begin{equation}
  \mathbf{o}_t = g \cdot \text{Attn}_t + (1 - g) \cdot \text{Decay}_t.
\end{equation}
After fusion, a learned output projection, residual addition, and layer normalization update the continuous stream. The feed-forward network (FFN) then applies
\begin{align}
 \mathbf{u}_t &= \operatorname{LN}_1\!\left(\mathbf{c}_t+W_O\mathbf{o}_t\right),\\
 \mathbf{b}_t &= \Theta(W_{\rm up}\mathbf{u}_t-\theta),\\
 \mathbf{c}'_t &= \operatorname{LN}_2\!\left(\mathbf{u}_t+W_{\rm down}\mathbf{b}_t\right),\qquad
 \mathbf{s}'_t=\Theta(\mathbf{c}'_t-\theta).
\end{align}
The two layer normalizations have distinct learned parameters, $W_{\rm up}\in\mathbb{R}^{D_{\rm ff}\times D}$, $W_{\rm down}\in\mathbb{R}^{D\times D_{\rm ff}}$, and $\theta=1$. Dropout is omitted from these inference equations. The FFN threshold is stateless, whereas the encoder LIF has membrane state. Both $\mathbf{s}'_t$ and $\mathbf{c}'_t$ pass to the next block. Only the down-projection consumes the binary FFN intermediate; the up-projection consumes $\mathbf{u}_t$.

\subsection{The ATan Surrogate Gradient}
\label{sec:atan}

The Heaviside step function $s = \Theta(u - \theta)$ is non-differentiable.
A scaled Sigmoid surrogate with $\alpha{=}10$ has peak derivative $\frac{\alpha}{4} = 2.5$, allowing worst-case local amplification above one in each layer.
The implemented backward rule uses an ATan-shaped surrogate:
\begin{equation}
 \frac{\partial s}{\partial u}\ \longleftarrow\ \widetilde g(u-\theta),\qquad
 \widetilde g(a)=\frac{1}{1+(\kappa a)^2},\quad\kappa=2.
\end{equation}
This assignment is not the analytic derivative of the hard step. Its peak magnitude is 1, which bounds this local surrogate derivative but does not bound the full network's gradient norm.

\subsection{Context MLP Head}
\label{sec:dynamic_prior}

The decoding head adds a bottleneck MLP on the continuous residual to the vocabulary projection:
\begin{equation}
  \mathbf{y}_t = W_{\text{vocab}} \mathbf{c}^{(L)}_t + 0.1 f_\phi(\mathbf{c}^{(L)}_t),
\end{equation}
where $f_\phi(\mathbf{c}) = W_2\operatorname{GELU}(W_1\mathbf{c})$ has hidden dimension $D/4 = 192$ and output dimension $V{=}48{,}000$ (${\sim}$9.4M parameters: $W_1 \in \mathbb{R}^{192 \times 768}$, $W_2 \in \mathbb{R}^{48000 \times 192}$).
Both MLP projections are bias-free. The head uses the last block\textquotesingle s already normalized output $\mathbf{c}^{(L)}_t$, with no additional final normalization. Applying softmax to $\mathbf{y}_t$ gives the next-token distribution. The factor 0.1 rescales the MLP logits; it does not bound their contribution relative to the linear branch.

\subsection{Model-Specific Bilingual Tokenizer}
\label{sec:tokenizer}

We replace the GPT-2 byte-level BPE tokenizer (50,257 entries, designed primarily for English) with a custom 48,000-token bilingual tokenizer trained via SentencePiece BPE~\citep{kudo2018sentencepiece} on a balanced Chinese--English corpus, with byte fallback for characters outside the learned pieces and extended subword merges (max piece length 24).
This 48K tokenizer is used for every 194M-scale model in this paper, including both GPT-2 baselines.
The 0.8B run uses a separate tokenizer of the same family (55,296 active entries in a 57,344-entry structural vocabulary); the two vocabularies are not interchangeable.
The extended merge length is motivated by sequential LIF computation: longer pieces can shorten $S$. There is no tokenizer ablation.

\subsection{Auxiliary Deep Supervision CE Training (AuxCE)}
\label{sec:auxce}

Each SymbolicLightBlock can produce exit logits via the shared decoding head.
During \emph{training}, cross-entropy losses at all block outputs provide deep supervision under our weighted objective:
\begin{equation}
  \mathcal{L}_{\text{total}} = \mathcal{L}_{\text{main}} + \frac{\lambda}{L} \sum_{i=0}^{L-1} \rho^{(L-1-i)} \cdot \mathcal{L}_{\text{exit}}^{(i)},
\end{equation}
where $\lambda{=}0.3$ is the auxiliary weight and $\rho{=}0.5$ exponentially down-weights shallower layers. The implementation divides the weighted exit-loss sum by $L$ before applying $\lambda$; this normalization is included above. The sum includes the final block: with the same shared head and targets, $\mathcal{L}_{\rm exit}^{(L-1)}=\mathcal{L}_{\rm main}$. The logged training loss is the main next-token cross-entropy, with the auxiliary loss logged separately.

\section{Training and Data Protocol}
\label{sec:setup}

\subsection{Model Configurations}

\begin{table}[H]
\centering
\caption{Primary model configurations. All 194M-scale models share the same SL-BPE 48K tokenizer. The 0.8B scale-up is reported separately in \Cref{sec:scaling_results} because it uses a different tokenizer and training mixture.}
\label{tab:config}
\small
\begin{adjustbox}{max width=\linewidth}
\begin{tabular}{@{}lccccccc@{}}
\toprule
\textbf{Model} & \textbf{$D$} & \textbf{Layers} & \textbf{Heads} & \textbf{$d_k$} & \textbf{$D_{\text{ff}}$} & \textbf{Vocab} & \textbf{Params} \\
\midrule
SymbolicLight V1 & 768 & 12 & 12 & 64 & 4,096 & 48,000 & 194M \\
GPT-2 124M & 768 & 12 & 12 & 64 & 3,072 & 48,000 & 122.7M \\
GPT-2 201M & 1,024 & 12 & 16 & 64 & 4,096 & 48,000 & 201M \\
\bottomrule
\end{tabular}
\end{adjustbox}
\end{table}

The GPT-2 201M baseline is a parameter-matched reimplementation of the GPT-2 architecture~\citep{radford2019language} ($d{=}1024$, $L{=}12$, $H{=}16$) and serves as the \textbf{primary comparison} throughout this paper.
Both GPT-2 baselines use the same SL-BPE 48K tokenizer and training corpus. The conventional label GPT-2 124M is retained, although its training log reports 122.7M parameters with this vocabulary.
The 124M variant matches the hybrid model's hidden dimension and depth, but the architectures allocate parameters differently, including the hybrid model's separate input embedding and output projection. The encoder leak and firing threshold are fixed constants, not learned LIF parameters.

A separate GPT-2 configuration ($d{=}1008$, $L{=}12$, $H{=}16$, 195.9M parameters) trained for 3B tokens on a rented 8$\times$RTX~5090 machine reached evaluation PPL 6.97; only a consolidated result record survives, so it is not a comparison anchor.

\paragraph{Choice of dense baseline.}
GPT-2 provides a full-precision Transformer reference under the same corpus, tokenizer, token budget, and hardware as the 194M SNN runs.
This comparison measures the quality difference between the complete spike-gated architecture and a conventional dense Transformer; it does not isolate activation binarization from the accompanying changes in sequence mixing and recurrent state.
Comparisons with dense linear-time or linear-attention models would answer a complementary efficiency question and are outside the controlled experiment reported here.

\paragraph{Parameter budget decomposition.}
\Cref{tab:param_budget} breaks down where the SNN's 194M parameters reside.
The model-specific components (decay path 7.3\%, context MLP 4.8\%) total ${\sim}$12\% of parameters, with the majority allocated to standard components shared by all language models.
The separate output projection (19.0\%) is architecturally standard but not weight-tied with the input embedding in this implementation; whether weight tying interacts favorably or unfavorably with sparse activations remains an open question for future work.

\begin{table}[H]
\centering
\caption{Parameter budget decomposition of the 194M SNN. Components marked with $^\star$ are model-specific; $^\ddagger$ marks the output projection, which is architecturally standard but not weight-tied with the input embedding in this implementation.}
\label{tab:param_budget}
\small
\begin{adjustbox}{max width=\linewidth}
\begin{tabular}{@{}lrr@{}}
\toprule
\textbf{Component} & \textbf{Params (M)} & \textbf{Share} \\
\midrule
Token Embedding & 36.9 & 19.0\% \\
Feed-Forward Network (FFN) & 75.5 & 38.9\% \\
Sparse Local Attention Path & 21.2 & 10.9\% \\
Output Projection$^\ddagger$ & 36.9 & 19.0\% \\
\midrule
Decay Path (TCAM projections)$^\star$ & 14.2 & 7.3\% \\
Context MLP Head$^\star$ & 9.4 & 4.8\% \\
Decay Factors + Fusion Gate$^\star$ & $<$0.1 & $<$0.1\% \\
LayerNorm (encoder + per-block) & $<$0.1 & $<$0.1\% \\
\midrule
\textbf{Total} & \textbf{194.0} & \textbf{100\%} \\
\bottomrule
\end{tabular}
\end{adjustbox}
\end{table}

\paragraph{Batch size note.}
The SNN and GPT-2 use different effective batch sizes (1,024 vs.\ 1,536) because SNN training requires more memory per sample due to membrane potential state.
Both models consume 3B tokens, but the resulting numbers of optimizer updates differ (5,722 vs.\ 3,814); the comparison therefore controls the token budget rather than the optimization trajectory.

\subsection{Training Protocol}

\begin{table}[H]
\centering
\caption{Training hyperparameters. All models consumed approximately 3B tokens using the same nominal source-sampling recipe. The 124M microbatch settings are taken from its console log; the 201M log establishes 786,432 tokens per update but does not establish the microbatch decomposition.}
\label{tab:training}
\small
\begin{adjustbox}{max width=\linewidth}
\begin{tabular}{@{}lcc@{}}
\toprule
\textbf{Hyperparameter} & \textbf{SNN} & \textbf{GPT-2} \\
\midrule
Total tokens & 3B & 3B \\
Optimizer & \multicolumn{2}{c}{AdamW ($\beta_1{=}0.9, \beta_2{=}0.95$)} \\
Peak learning rate & $3 \times 10^{-4}$ & $3 \times 10^{-4}$ \\
LR schedule & \multicolumn{2}{c}{Cosine annealing + linear warmup (2,000 steps)} \\
Weight decay & 0.1 & 0.1 \\
Gradient clipping & 1.0 & 1.0 \\
Precision & BF16 & BF16 \\
Per-GPU batch size & 16 & 64 (124M); unverified (201M) \\
Gradient accumulation & 16 & 6 (124M); unverified (201M) \\
Effective batch size & 1,024 & 1,536 \\
Sequence length & 512 & 512 \\
GPUs & 4$\times$ A800 & 4$\times$ A800 \\
\bottomrule
\end{tabular}
\end{adjustbox}
\end{table}

\subsection{Training Data}

\begin{table}[H]
\centering
\caption{Nominal document-sampling probabilities by domain. A source is selected before tokenizing a complete document, so these probabilities are not measured token fractions. All 194M-scale models use this recipe; raw text is not redistributed.}
\label{tab:data}
\small
\begin{adjustbox}{max width=\linewidth}
\begin{tabular}{@{}llc@{}}
\toprule
\textbf{Category} & \textbf{Domain profile} & \textbf{Sampling probability} \\
\midrule
\multirow{3}{*}{Chinese (40\%)} & Chinese-Reference & 15\% \\
& Chinese-Web & 15\% \\
& Chinese-General & 10\% \\
\midrule
\multirow{4}{*}{English (40\%)} & English-Educational & 10\% \\
& English-Math & 10\% \\
& English-Reference & 10\% \\
& Math-Web & 10\% \\
\midrule
Code (5\%) & Code & 5\% \\
\midrule
\multirow{2}{*}{Narrative (15\%)} & English-Narrative & 5\% \\
& Chinese-Narrative & 10\% \\
\bottomrule
\end{tabular}
\end{adjustbox}
\end{table}

The source streams have heterogeneous redistribution terms, so raw text and shard records are not redistributed. The recipe selects a source for each document, then appends the entire tokenized document to the training buffer. Consequently, differences in document length can change the realized token mixture. The four Chinese domains have a combined nominal document-sampling probability of 50\%, including Chinese-Narrative. All 194M-scale comparisons use this recipe, tokenizer, and last-200 evaluation procedure.

\subsection{Evaluation Protocol}

All reported 194M PPL values use the last 200 documents of each of the ten local domain streams. The evaluation loader sorts the source files, concatenates each domain's selected texts without inserting document separators, and scores non-overlapping 512-token target chunks, discarding the incomplete tail. Each chunk is evaluated independently; recurrent state is not carried between chunks. Token counts in \Cref{tab:perdomain} refer to scored targets.

\paragraph{Training-set separation.}
The available training loaders read the matching source files without an explicit last-200 exclusion, and some training logs report exhausted sources being restarted. No verified pre-training split manifest or train/evaluation overlap audit is available. A prior physical split could change the overlap status, but it has not been established. We therefore describe these values as fixed-set evaluation PPL, not as verified unseen-data performance; shared evaluation text alone does not remove possible contamination bias.

Overall PPL is the exponential of the token-weighted mean cross-entropy,
\begin{equation}\text{PPL}=\exp\!\left(\frac{\sum_{d,j}T_{d,j}\mathcal{L}_{d,j}}{\sum_{d,j}T_{d,j}}\right).\end{equation}
Here $d$ indexes domains, $j$ indexes scored chunks, $T_{d,j}$ counts target tokens, and $\mathcal{L}_{d,j}$ is their mean next-token cross-entropy using natural logarithms.
Code accounts for 43.7\% of the 2,984,960 scored tokens, while the four Chinese domains together account for 9.5\%. Their nominal document-sampling probabilities are 5\% and 50\%, respectively; these are not observed training-token fractions.
The four-run token-weighted SNN mean is 8.904, and the arithmetic mean of the ten domain PPLs is 29.38. Weighting domain cross-entropies by the nominal probabilities in \Cref{tab:data} gives PPL 22.4. This last number is a recipe-reweighted diagnostic, not a reconstruction of the realized training-token distribution.

\section{Experiments and Results}
\label{sec:results}

\subsection{Main Results}

\begin{table}[H]
\centering
\caption{Token-weighted evaluation PPL on the last-200-sample shards.
Four independent SNN runs (two with AuxCE, two without) reach PPL 8.88--8.93 on 4$\times$ A800-40GB with the SL-BPE 48K tokenizer.
The SNN mean is 7.7\% above GPT-2 201M (PPL 8.27) and 0.06 below the single GPT-2 124M run (PPL 8.96). Encoder zero-spike fractions are training-probe summaries; separation of the evaluation set from training is unverified.}
\label{tab:main}
\small
\begin{adjustbox}{max width=\linewidth}
\begin{tabular}{@{}llcccc@{}}
\toprule
\textbf{Model} & \textbf{Params} & \textbf{Training} & \textbf{Seed} & \textbf{Eval PPL $\downarrow$} & \textbf{Encoder probes} \\
\midrule
SNN (AuxCE) & 194M & AuxCE & 123 & 8.91 & $>$89\% \\
SNN (AuxCE) & 194M & AuxCE & 456 & 8.88 & $>$89\% \\
SNN (noAuxCE) & 194M & noAuxCE & 42 & 8.90 & $>$89\% \\
SNN (noAuxCE) & 194M & noAuxCE & 123 & 8.93 & $>$89\% \\
\cmidrule{1-6}
\multicolumn{2}{@{}l}{\textbf{SNN mean $\pm$ sample std}} & & & \textbf{8.904 $\pm$ 0.019} & $>$89\% \\
\midrule
GPT-2 & 122.7M & Standard & --- & 8.96 & --- \\
GPT-2 & 201M & Standard & --- & \textbf{8.27} & --- \\
\bottomrule
\end{tabular}
\end{adjustbox}
\end{table}

The four SNN runs span two training objectives (AuxCE vs.\ noAuxCE) and three random seeds (seed~123 is shared across objectives).
From the unrounded evaluation outputs, the token-weighted mean PPL is 8.904 with sample standard deviation 0.019.
That mean is 7.7\% above the single GPT-2 201M run and 0.059 below the single GPT-2 124M run; each dense baseline was trained once.

\subsection{Per-Domain Analysis}

\begin{table}[H]
\centering
\caption{Per-domain evaluation PPL with scored token counts. SNN values are four-run means; GPT-2 baselines are single A800 runs. Overall is token-weighted. Shares use the shared 2,984,960-token shard.}
\label{tab:perdomain}
\small
\setlength{\tabcolsep}{4pt}
\begin{adjustbox}{max width=\linewidth}
\begin{tabular}{@{}lrrccc@{}}
\toprule
\textbf{Domain} & \textbf{Tokens} & \textbf{Share} & \textbf{SNN} & \textbf{GPT-2 124M} & \textbf{GPT-2 201M} \\
\midrule
Chinese-Reference & 27,648 & 0.9\% & 12.53 & 10.49 & 9.47 \\
Chinese-Web       & 141,824 & 4.8\% & 101.71 & 109.65 & 95.35 \\
Chinese-General   & 93,184 & 3.1\% & 63.65 & 67.70 & 59.70 \\
English-Educational & 232,960 & 7.8\% & 10.49 & 10.91 & 9.66 \\
English-Math      & 222,720 & 7.5\% & 14.66 & 15.12 & 13.36 \\
English-Reference & 211,968 & 7.1\% & 19.29 & 19.58 & 17.69 \\
Math-Web          & 687,104 & 23.0\% & 14.28 & 14.27 & 13.11 \\
Code              & 1,303,552 & 43.7\% & 3.51 & 3.46 & 3.30 \\
English-Narrative & 41,984 & 1.4\% & 9.99 & 10.47 & 9.36 \\
Chinese-Narrative & 22,016 & 0.7\% & 43.69 & 48.96 & 42.21 \\
\midrule
\textbf{Token-weighted} & 2,984,960 & 100\% & 8.904 & 8.96 & \textbf{8.27} \\
Unweighted domain mean & --- & --- & 29.38 & 31.07 & 27.32 \\
\bottomrule
\end{tabular}
\end{adjustbox}
\end{table}

The SNN has lower PPL than GPT-2 124M in 7 of 10 domains, with the largest relative reductions on Chinese-Narrative, Chinese-Web, and Chinese-General.
It trails on Chinese-Reference ($+19.4\%$) and is nearly tied on Code and Math-Web, the two domains that together are 66.7\% of evaluation tokens.
The unweighted domain mean (29.38) and the recipe-reweighted PPL (22.4) sit well above the token-weighted headline of 8.904 because Code (PPL 3.51) dominates token mass.

\subsection{AuxCE vs.\ noAuxCE}

\begin{table}[H]
\centering
\caption{AuxCE vs.\ noAuxCE comparison. Both achieve nearly identical evaluation quality.}
\label{tab:auxce}
\small
\begin{adjustbox}{max width=\linewidth}
\begin{tabular}{@{}lcccc@{}}
\toprule
\textbf{Variant} & \textbf{Seeds} & \textbf{Mean Eval PPL} & \textbf{Steps} & \textbf{Optimization evidence} \\
\midrule
AuxCE ($\lambda{=}0.3$) & 123, 456 & 8.897 & 5,722 & No matched-time benefit established \\
noAuxCE & 42, 123 & 8.912 & 5,722 & Baseline \\
\midrule
\textbf{Difference} & & \textbf{$\Delta$ = 0.015} & & (0.17\%) \\
\bottomrule
\end{tabular}
\end{adjustbox}
\end{table}

AuxCE and noAuxCE achieve virtually identical evaluation quality ($\Delta < 0.3\%$).
The evaluation values do not establish a quality difference between objectives. Training traces use different seeds and fluctuate with sampled data; the longer AuxCE wall-clock times in \Cref{tab:compute} do not establish a time-to-quality advantage.

\subsection{Component Ablation}
\label{sec:ablation}

\begin{table}[H]
\centering
\caption{Component ablation at 0.5B tokens, same data, tokenizer, sequence length, and optimizer.
The three removal variants were trained with AuxCE on 4$\times$A800; the 0.5B full-model reference used noAuxCE on 8$\times$RTX~5090.}
\label{tab:ablation}
\small
\begin{adjustbox}{max width=\linewidth}
\begin{tabular}{@{}lcccl@{}}
\toprule
\textbf{Variant} & \textbf{Params} & \textbf{Eval PPL} & \textbf{$\Delta$ PPL} & \textbf{Modification} \\
\midrule
Full model (0.5B) & 194M & 17.72 & --- & (reference, 0.5B tokens) \\
\midrule
Static Prior & 185M & 21.26 & $+20.0\%$ & Context MLP $\to$ static $\log\pi$ \\
Decay Only & 164M & 25.27 & $+42.6\%$ & Local attention \emph{and} context MLP removed \\
No Attention & 173M & 38.56 & $+117.6\%$ & SparseLocalAttention path removed \\
\midrule
\textbf{Full model (3B)} & \textbf{194M} & \textbf{8.91} & \textbf{---} & \textbf{(converged reference, 3B tokens)} \\
\bottomrule
\end{tabular}
\end{adjustbox}
\end{table}

All ablation variants and the 0.5B full-model reference consume 0.5B training tokens on the same data, tokenizer, sequence length, learning-rate schedule, and optimizer.
The three removals were trained with AuxCE on 4$\times$A800; the 0.5B full-model reference used noAuxCE on 8$\times$RTX~5090, so the $\Delta$ values share a token budget but not hardware or objective.
``No Attention'' (decay path + context MLP, PPL 38.56) is the largest gap versus the 0.5B full model (17.72), and it is worse than ``Decay Only'' (25.27), but these cross-configuration differences do not identify individual component effects or their interactions.
The 3B full-model PPL 8.91 is shown for context and is not the $\Delta$ baseline. Causal attribution would require a full-model control with the same objective and optimization settings at 0.5B tokens; final 3B AuxCE similarity does not supply that control.
Per-domain values are in \Cref{app:ablation_perdomain}.

\subsection{Training Dynamics and Sparsity}
\label{sec:training_dynamics}

\begin{figure}[htbp]
\centering
\includegraphics[width=\linewidth]{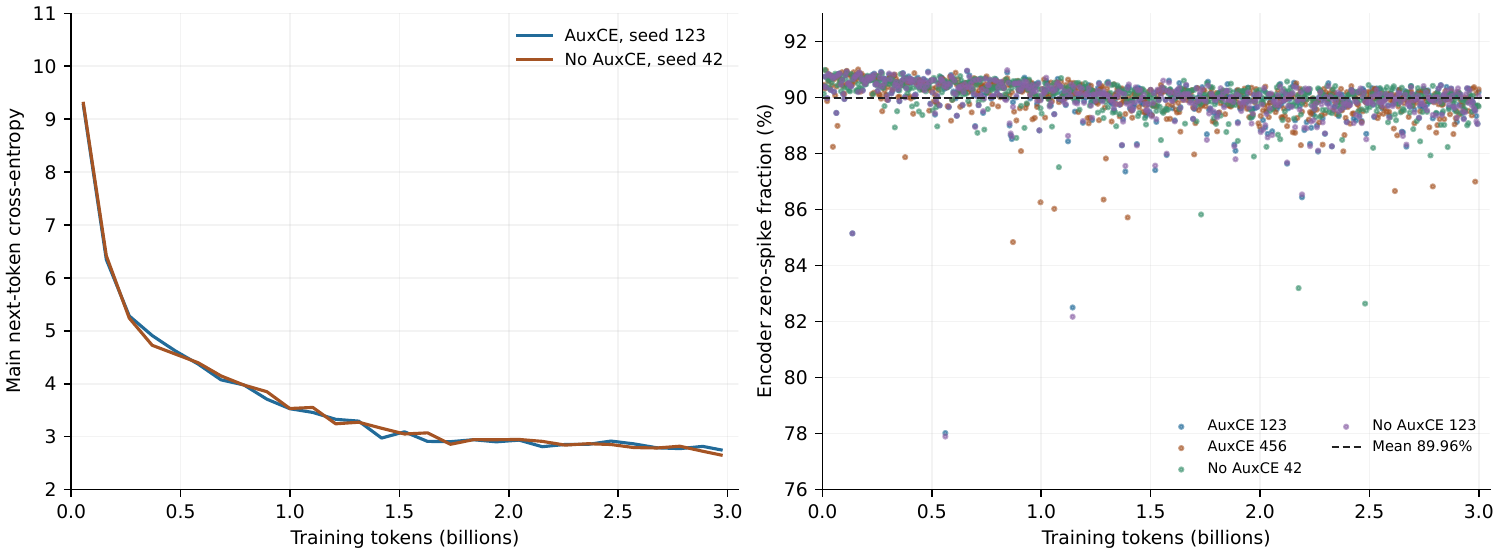}
\caption{\textbf{Left:} Main next-token training cross-entropy versus tokens consumed, binned over 20 consecutive logged observations for the indicated objective and seed. \textbf{Right:} All 2,288 encoder probes from four runs; each probes $32\times768$ output elements. The dashed line is the 89.96\% mean across probes. The plotted range includes the 77.9\% minimum. These are training diagnostics, not evaluation-set sparsity measurements.}
\label{fig:training}
\end{figure}

\paragraph{Convergence and encoder probes.}
\Cref{fig:training} is regenerated from the step-level records. Both curves report main next-token cross-entropy; AuxCE's auxiliary loss is recorded separately. For legibility the left panel groups successive logged observations into non-overlapping bins of 20 and plots each bin's mean token count and mean loss, retaining the final partial bin. Raw final losses are 2.8721 (AuxCE, seed 123) and 2.3453 (noAuxCE, seed 42). These differ from fixed-set PPL because the data and aggregation differ; the logged AuxCE loss is not inflated by adding the auxiliary term.

Every ten optimizer steps, the logger separately runs the encoder on the first 32 tokens of the first sample of the current microbatch. Across four runs this gives 2,288 probes. Their mean zero-spike fraction is 89.96\%, the per-run means range from 89.94\% to 90.01\%, 94.7\% of probes are at least 89\%, and the minimum is 77.9\%. This measures sampled encoder outputs, not full-batch or whole-model activation sparsity. The right panel plots every recorded probe with an uncropped vertical range and a mean line; no uncertainty band is implied.

\paragraph{ATan vs.\ Sigmoid surrogate.}
A 2,000-step comparison (\Cref{fig:surrogate}) reveals that ATan produces substantially stronger gradient signals: mean pre-clip gradient norm 2.94 (vs.\ 0.94 for Sigmoid), with ATan often exceeding the gradient-clipping threshold of 1.0, while the Sigmoid mean is just below that threshold and some Sigmoid observations exceed it.
Both surrogates reach similar final loss after 2,000 steps (ATan: 2.62, Sigmoid: 2.55).
These figures come from a retained log of 201 points sampled every 10 steps; the means are taken over all logged points (peak pre-clip norms 12.36 for ATan and 2.43 for Sigmoid), and the final-loss values are trailing averages over the last 10 logged points, the single last points being 2.75 and 2.67.
This comparison predates the 194M bilingual configuration: it was run under the earlier prototype setup, on a single-domain English subset with the 50,257-entry GPT-2 tokenizer and batch size 8.
It therefore supports the surrogate choice as directional evidence and should not be read as a measurement taken on the 3B-token bilingual runs reported above.
ATan's peak derivative is bounded by 1.0, whereas the scaled Sigmoid permits a local derivative as large as 2.5 before gradient clipping.

\begin{figure}[htbp]
\centering
\begin{tikzpicture}
\begin{axis}[
    width=0.65\textwidth, height=4.5cm,
    xlabel={Step}, ylabel={Pre-clip Gradient Norm},
    xmin=0, xmax=2050, ymin=0, ymax=14,
    grid=major, grid style={gray!20},
    legend pos=north east, legend style={font=\scriptsize},
]
\addplot[spikeblue, thick, mark=none, smooth]
    coordinates {(1,2.09)(20,2.21)(40,2.44)(60,1.54)(80,1.04)(100,1.08)(140,1.59)(200,1.45)(300,1.24)(400,1.41)(500,1.88)(600,1.91)(640,12.36)(700,2.49)(800,3.03)(900,2.19)(1000,3.69)(1100,2.80)(1200,2.76)(1300,4.79)(1400,3.85)(1500,2.95)(1600,4.92)(1700,3.97)(1800,3.52)(1900,4.36)(2000,5.50)};
\addlegendentry{ATan (mean 2.94)}
\addplot[spikered, thick, mark=none, smooth]
    coordinates {(1,2.26)(20,2.10)(40,2.25)(60,1.46)(80,0.91)(100,0.85)(140,1.00)(200,1.18)(300,0.82)(400,0.86)(500,0.95)(600,0.90)(700,0.91)(800,0.85)(900,0.87)(1000,0.91)(1100,0.82)(1200,0.90)(1300,0.89)(1400,0.94)(1500,0.87)(1600,0.92)(1700,0.91)(1800,0.86)(1900,0.96)(2000,0.94)};
\addlegendentry{Sigmoid (mean 0.94)}
\draw[gray, densely dashed] (axis cs:0,1.0) -- (axis cs:2050,1.0) node[right, font=\tiny] {clip};
\end{axis}
\end{tikzpicture}
\caption{Pre-clip gradient norms over 2,000 prototype-training steps. ATan has a larger mean norm than Sigmoid and more often exceeds the clipping threshold.}
\label{fig:surrogate}
\end{figure}

\paragraph{Learned gate values.}
The fusion gate $g = \sigma(w_g)$ between the attention and decay paths converges to similar depth-wise profiles across the four runs (\Cref{tab:gates}).
Early layers ($0$--$3$) show near-equal mixing ($g \approx 0.50$--$0.54$), while deeper layers ($7$--$11$) shift toward attention dominance ($g \approx 0.58$--$0.63$).
The learned coefficient is generally larger in deeper blocks. It is not an attention contribution fraction: interpreting branch importance also requires their output magnitudes, which are not measured here.
Across-run population standard deviations range from 0.0007 to 0.0264 across layers; most layers are below 0.02, with the largest dispersion at Block~10.
The full 12-layer, 4-run gate and decay factor table is provided in \Cref{app:gate_full}.

\begin{table}[H]
\centering
\caption{Learned gate values $g = \sigma(w_g)$ across four runs spanning three seeds and two objectives. Higher $g$ = more attention, lower $g$ = more decay.}
\label{tab:gates}
\small
\begin{adjustbox}{max width=\linewidth}
\begin{tabular}{@{}lcccccc@{}}
\toprule
\textbf{Block} & \textbf{AuxCE 123} & \textbf{AuxCE 456} & \textbf{noAuxCE 42} & \textbf{noAuxCE 123} & \textbf{Mean} & \textbf{Interp.} \\
\midrule
0  & 0.490 & 0.489 & 0.489 & 0.488 & 0.489 & Balanced \\
3  & 0.537 & 0.527 & 0.547 & 0.533 & 0.536 & Balanced \\
6  & 0.578 & 0.557 & 0.576 & 0.573 & 0.571 & Attn-leaning \\
9  & 0.564 & 0.604 & 0.596 & 0.561 & 0.582 & Attn-leaning \\
11 & 0.609 & 0.579 & 0.563 & 0.582 & 0.583 & Attn-leaning \\
\midrule
\textbf{All-layer mean} & 0.563 & 0.562 & 0.560 & 0.559 & \textbf{0.561} & Gate mean \\
\bottomrule
\end{tabular}
\end{adjustbox}
\end{table}

\paragraph{Decay factors.}
The per-head exponential decay factors $\alpha_h = \sigma(\gamma_h)$ show an overall increase from ${\sim}0.91$ in Block~0 to ${\sim}0.95$ in Block~11, with a small local decrease from Block~1 to Block~2.
This trend indicates slower direct state decay in deeper layers overall. Pooling heads across the four runs gives half-lives of about 7--12 tokens in Blocks 0--5 and 9--15 in Blocks 6--11 (\Cref{app:perhead_decay}).

\Cref{fig:gate_decay} visualizes both the gate and decay trends across layers.

\begin{figure}[htbp]
\centering
\begin{tikzpicture}
\begin{axis}[
    width=0.48\textwidth, height=4.5cm,
    xlabel={Block}, ylabel={Mixing coefficient $g$},
    xmin=-0.5, xmax=11.5, ymin=0.45, ymax=0.65,
    xtick={0,1,...,11},
    grid=major, grid style={gray!20},
    title style={font=\small\bfseries},
    legend pos=south east, legend style={font=\scriptsize},
]
\addplot[spikeblue, thick, mark=*, mark size=1.5pt, error bars/.cd, y dir=both, y explicit]
    coordinates {
    (0,0.4892) +- (0,0.0007)
    (1,0.5255) +- (0,0.0057)
    (2,0.5315) +- (0,0.0076)
    (3,0.5360) +- (0,0.0071)
    (4,0.5553) +- (0,0.0097)
    (5,0.5759) +- (0,0.0093)
    (6,0.5711) +- (0,0.0082)
    (7,0.5929) +- (0,0.0153)
    (8,0.5895) +- (0,0.0100)
    (9,0.5815) +- (0,0.0189)
    (10,0.6016) +- (0,0.0264)
    (11,0.5831) +- (0,0.0165)
    };
\addlegendentry{4-run mean $\pm\sigma$}
\draw[gray, densely dashed] (axis cs:-0.5,0.5) -- (axis cs:11.5,0.5);
\node[font=\tiny, gray] at (axis cs:10,0.47) {50/50};
\end{axis}
\end{tikzpicture}%
\hfill
\begin{tikzpicture}
\begin{axis}[
    width=0.48\textwidth, height=4.5cm,
    xlabel={Block}, ylabel={Decay $\alpha$},
    xmin=-0.5, xmax=11.5, ymin=0.910, ymax=0.955,
    xtick={0,1,...,11},
    grid=major, grid style={gray!20},
    title style={font=\small\bfseries},
]
\addplot[spikepurple, thick, mark=square*, mark size=1.5pt]
    coordinates {(0,0.9149)(1,0.9291)(2,0.9271)(3,0.9306)(4,0.9332)(5,0.9358)(6,0.9381)(7,0.9414)(8,0.9432)(9,0.9437)(10,0.9482)(11,0.9488)};
\end{axis}
\end{tikzpicture}
\caption{\textbf{Left:} Learned gate values across layers (4-run mean $\pm \sigma$). The profile shifts from balanced decay/attention mixing in shallow layers toward greater attention weight in deeper layers. \textbf{Right:} Exponential decay factors generally increase with depth, with a small Block~1-to-2 reversal.}
\label{fig:gate_decay}
\end{figure}

\subsection{Scaling Results: 0.8B SymbolicLight V1}
\label{sec:scaling_results}

To test whether the training stack extends beyond 200M parameters, we trained a 0.8B-parameter SymbolicLight V1 checkpoint for 48.8B tokens.
The run uses a larger tokenizer, $w{=}512$ (full training context), and a separate 14-source mixture, and has no matched dense baseline or post-training alignment.

\begin{table}[H]
\centering
\caption{Configuration and evaluation diagnostics of the 0.8B scale-up run. All three PPL values use the same 186,000-step checkpoint.}
\label{tab:scale08b}
\small
\setlength{\tabcolsep}{3pt}
\begin{adjustbox}{max width=\linewidth}
\begin{tabular}{@{}p{0.22\textwidth}p{0.60\textwidth}@{}}
\toprule
\textbf{Field} & \textbf{0.8B SymbolicLight V1 status} \\
\midrule
Parameter count & 873,668,135 parameters \\
Model configuration & $d{=}1536$, 22 layers, 24 heads, $D_{\text{ff}}{=}6144$, sliding-window width 512 (full training context) \\
Tokenizer & 55,296 active entries within a 57,344-entry structural vocabulary \\
Pre-training mixture & Separate 14-source mixture \\
Training tokens & 48.8B tokens (186,000 steps $\times$ 262,144 tokens per step) \\
Context length & 512 tokens \\
Evaluation PPL & 14.17 on a 2,015,232-token evaluation sequence; 16.79 on its first 131,072 tokens; 22.53 on a separate 65,536-token evaluation excerpt \\
Activation sparsity & 95.44\% at the encoder and 93.77\% mean across blocks in an eight-prompt inference audit \\
Post-training alignment & None \\
Public artifact & Checkpoint, tokenizer assets, loader, and generation entry point \\
\bottomrule
\end{tabular}
\end{adjustbox}
\end{table}

The three PPL figures are protocol-sensitive diagnostics on one checkpoint, not a comparison with the 194M tables or with a dense model.
The public artifact loads and generates with the released inference code.

\subsection{Zero-Shot Downstream Tasks}
\label{sec:zeroshot}

To assess whether the SNN's representations support reasoning beyond language modeling, we evaluate on five standard zero-shot benchmarks (\Cref{tab:zeroshot}): PIQA~\citep{bisk2020piqa} (physical commonsense), SciQ~\citep{welbl2017sciq} (science QA), LAMBADA~\citep{paperno2016lambada} (long-range word prediction), HellaSwag~\citep{zellers2019hellaswag} (commonsense sentence completion), and ARC-Easy~\citep{clark2018arc} (grade-school science).

\begin{table}[H]
\centering
\caption{Zero-shot accuracy on full test/validation splits with 95\% bootstrap confidence intervals (10,000 resamples). LAMBADA is scored on the final subword token, not the standard whole-word protocol.}
\label{tab:zeroshot}
\small
\begin{adjustbox}{max width=\linewidth}
\begin{tabular}{@{}lccccc@{}}
\toprule
\textbf{Model} & \textbf{PIQA} & \textbf{SciQ} & \textbf{LAMBADA} & \textbf{HellaSwag} & \textbf{ARC-E} \\
 & \textit{n=1838} & \textit{n=1000} & \textit{n=5153} & \textit{n=10042} & \textit{n=2376} \\
\midrule
Random & 50.0\% & 25.0\% & ${\sim}$0\% & 25.0\% & 25.0\% \\
\midrule
SNN-194M & 53.8 {\tiny[51.5,56.1]} & 31.1 {\tiny[28.2,34.0]} & 80.2 {\tiny[79.1,81.3]} & 26.5 {\tiny[25.6,27.4]} & \textbf{32.1} {\tiny[30.2,34.0]} \\
GPT2-201M & \textbf{54.9} {\tiny[52.7,57.2]} & \textbf{31.9} {\tiny[29.1,34.8]} & \textbf{80.6} {\tiny[79.5,81.7]} & \textbf{26.7} {\tiny[25.8,27.5]} & 31.9 {\tiny[30.1,33.8]} \\
\bottomrule
\end{tabular}
\end{adjustbox}
\end{table}

All five benchmarks use their full test or validation splits (20,409 examples in total).
The absolute SNN--GPT-2 differences range from 0.2 to 1.1 percentage points, and the separately reported bootstrap intervals overlap on every task. Interval overlap is not a paired test of the model difference or evidence of equivalence; per-example outcomes were not retained to estimate paired difference intervals.
Both models remain close to chance on PIQA and HellaSwag.

\paragraph{LAMBADA scoring protocol.}
We score LAMBADA on the final \emph{sub-word token}: the model receives the passage minus that token and must reproduce it by greedy argmax.
For multi-piece target words, the leading pieces remain in the context, making the scored prediction easier than requiring the complete word. The reported $>$80\% therefore cannot establish standard whole-word accuracy.
Both models are scored identically, so the descriptive difference is 0.4 percentage points, but the absolute values belong to this protocol and should not be read against published LAMBADA numbers.
The current zero-shot suite is English-only despite the bilingual pre-training mixture; Chinese task evaluation remains outside the present evidence.

\subsection{Generation Quality}
\label{sec:generation}

We compare free-form text generation from the SNN (AuxCE seed 123) and GPT-2 201M using 10 prompts spanning English narrative, science, mathematics, Chinese text, and Python code, under sampling decoding (temperature 0.7, top-$k$ 50).

\begin{table}[H]
\centering
\caption{4-gram repetition rate in generated text (lower is better). Sampling decoding, 150 tokens per prompt. The SNN produces less repetitive text on 8 of 10 prompts.}
\label{tab:generation}
\small
\begin{adjustbox}{max width=\linewidth}
\begin{tabular}{@{}lccc@{}}
\toprule
\textbf{Domain} & \textbf{SNN-194M} & \textbf{GPT2-201M} & \textbf{Winner} \\
\midrule
EN-Story      & \textbf{6.1\%} & 14.3\% & SNN \\
EN-Science    & \textbf{26.5\%} & 63.3\% & SNN \\
EN-Academic   & \textbf{25.2\%} & 50.3\% & SNN \\
EN-Physics    & 38.8\% & \textbf{28.6\%} & GPT-2 \\
ZH-Story      & \textbf{16.3\%} & 64.6\% & SNN \\
ZH-Knowledge  & \textbf{31.3\%} & 45.6\% & SNN \\
ZH-Tech       & \textbf{5.4\%} & 49.7\% & SNN \\
Code-General  & \textbf{50.3\%} & 59.9\% & SNN \\
Code-Numpy    & \textbf{32.0\%} & 91.8\% & SNN \\
EN-Math       & 41.5\% & \textbf{24.5\%} & GPT-2 \\
\midrule
\textbf{Mean} & \textbf{27.3\%} & 49.3\% & SNN \\
\bottomrule
\end{tabular}
\end{adjustbox}
\end{table}

The SNN produces lower repetition rates (mean 27.3\% vs.\ 49.3\%), winning 8 of 10 prompts.
A decoder-sensitivity analysis tests whether this ordering persists when sampling is removed or adapted.

\paragraph{Decoder sensitivity.}
\label{sec:greedy_controlled}
We re-run the 10 prompts under greedy decoding and under an entropy-modulated sampling rule.
Greedy decoding removes sampling randomness but does not equalize the entropy or shape of the two models' output distributions.
\Cref{tab:decoding_controlled} compares all three decoding configurations.

\begin{table}[H]
\centering
\caption{Mean 4-gram repetition rate over 10 prompts under three decoding strategies. The ranking changes with the decoding rule. The entropy-to-sampling transformation is not specified in the retained aggregate record, so the adaptive row is a historical diagnostic rather than a reproducible decoder comparison.}
\label{tab:decoding_controlled}
\small
\begin{adjustbox}{max width=\linewidth}
\begin{tabular}{@{}lccc@{}}
\toprule
\textbf{Decoding} & \textbf{SNN-194M} & \textbf{GPT2-201M} & \textbf{$\Delta$ (SNN $-$ GPT-2)} \\
\midrule
Greedy (top-$k{=}1$)            & 71.16\% & 78.09\% & $-6.93$~pp \\
Sampling ($T{=}0.7$, top-$k{=}50$) & 27.35\% & 49.25\% & $-21.90$~pp \\
Adaptive ($T{=}0.8$, top-$k{=}50$, entropy-modulated) & 36.19\% & 25.37\% & $+10.82$~pp \\
\bottomrule
\end{tabular}
\end{adjustbox}
\end{table}

The SNN retains a 6.9-point repetition advantage under greedy decoding, but the adaptive rule reverses the ordering by 10.8 points.
Because decoding changes both the selected tokens and the states subsequently presented to each autoregressive model, these results do not decompose repetition into architecture and entropy components.
The ten-prompt records do not specify whether 4-grams are formed from tokens, words, or characters, nor the exact entropy-modulation function. We retain their reported rates as historical summaries rather than treating them as directly comparable to the fully specified 500-text lexical metric.

\paragraph{Large-scale lexical diversity.}
We next generate 500 samples per model (128 tokens each, 20 English prompts, temperature 0.7, top-$k$ 50) and compute Distinct-$n$ metrics~(\Cref{tab:distinct}).

\begin{table}[H]
\centering
\caption{Distinct-$n$ (higher is better) and 4-gram repetition rate (lower is better) over 500 generated samples per model (temperature 0.7, top-$k$ 50, 128 tokens). The SNN achieves higher lexical diversity across all $n$-gram orders.}
\label{tab:distinct}
\small
\begin{adjustbox}{max width=\linewidth}
\begin{tabular}{@{}lcccccl@{}}
\toprule
\textbf{Model} & \textbf{Dist-1} & \textbf{Dist-2} & \textbf{Dist-3} & \textbf{Dist-4} & \textbf{Rep-4} & \textbf{Winner} \\
\midrule
SNN-194M  & \textbf{0.172} & \textbf{0.491} & \textbf{0.737} & \textbf{0.856} & \textbf{7.7\%} & \multirow{2}{*}{SNN (all $n$)} \\
GPT2-201M & 0.150 & 0.438 & 0.653 & 0.760 & 16.8\% & \\
\bottomrule
\end{tabular}
\end{adjustbox}
\end{table}

The script splits decoded text on whitespace and includes the prompt along with its 128 generated tokens. For Distinct-$n$, all within-text $n$-grams are pooled across 500 texts and the number of unique $n$-grams is divided by their total count. Repetition counts duplicate 4-grams within each text, then sums duplicate counts and denominators across texts. The 20 prompts are all English, each used 25 times. These definitions and prompt inclusion differ from a continuation-only bilingual evaluation; no sampling seed or generated-text archive is available for an exact rerun.
The SNN has higher Distinct-$n$ across all reported orders: Dist-1 is 14.7\% higher, Dist-2 is 12.1\% higher, Dist-3 is 12.9\% higher, and Dist-4 is 12.6\% higher, while the 4-gram repetition rate is 54\% lower (7.7\% vs.\ 16.8\%).
This larger sample corroborates the lower-repetition result for the specified sampling configuration and the GPT-2 201M baseline.

The retained narrative excerpts contain sentence-level structure but also grammatical, referential, and semantic errors. Scientific prose is not factually reliable; these observations do not establish a general capability claim.
Code generation is a shared weakness. The nominal 5\% code document-sampling probability does not establish its realized training-token share or explain this weakness.
On the Chinese prompts in the ten-prompt set, SNN generation has lower reported repetition. Both models use the same bilingual tokenizer, so these results do not identify a tokenizer explanation for their difference.
Representative generation samples are provided in \Cref{app:generation}.

Generation speed on GPU is approximately 22.8 tok/s for the SNN vs.\ 91.5 tok/s for GPT-2 (see \Cref{sec:inference_bench} for detailed benchmarks).

\subsection{Inference Benchmark: GPU vs.\ CPU}
\label{sec:inference_bench}

To characterize the SNN's performance profile on consumer-grade hardware representative of edge deployment scenarios, we benchmark autoregressive generation on an NVIDIA RTX 2080 Ti GPU and an AMD Ryzen 7 5800X CPU (\Cref{tab:inference}).

\begin{table}[H]
\centering
\caption{Inference benchmark (200 tokens generated, temperature 0.7, top-$k$ 50, 5-run average after 3 warmup runs). GPU power is queried once after each timed generation via \texttt{nvidia-smi}, so GPU energy is a rough power-times-latency estimate, not time-integrated generation energy. CPU values use LibreHardwareMonitor package-power records. The scope is device/package power, not whole-system energy.}
\label{tab:inference}
\small
\begin{adjustbox}{max width=\linewidth}
\begin{tabular}{@{}llcccc@{}}
\toprule
\textbf{Model} & \textbf{Device} & \textbf{tok/s} & \textbf{ms/tok} & \textbf{Power (W)} & \textbf{Energy estimate (mJ/tok)} \\
\midrule
SNN-194M  & GPU (RTX 2080 Ti) & 22.8 & 43.8 & 65 & 2,848 \\
GPT2-201M & GPU (RTX 2080 Ti) & 91.5 & 10.9 & 83 & 905 \\
\midrule
SNN-194M  & CPU (Ryzen 7 5800X) & 19.8 & 50.5 & 90 & 4,559 \\
GPT2-201M & CPU (Ryzen 7 5800X) & 24.6 & 40.7 & 92 & 3,752 \\
\bottomrule
\end{tabular}
\end{adjustbox}
\end{table}

Three findings emerge:

\paragraph{GPU: GPT-2 dominates.}
On GPU, GPT-2 is $4.0\times$ faster than the SNN (91.5 vs.\ 22.8 tok/s), reflecting the GPU's optimization for dense matrix operations.
The SNN's LIF membrane potential updates are sequential and do not map to dense GPU kernels as GPT-2 does.
The post-generation GPU power reading is 22\% lower for the SNN (65W vs.\ 83W), which does not offset the $4\times$ latency gap, so the power-times-latency estimate is $3.1\times$ higher (2,848 vs.\ 905\,mJ/tok).

\paragraph{CPU: narrower speed gap, comparable power.}
On CPU, the speed gap narrows to $1.24\times$: the SNN achieves $80\%$ of GPT-2's throughput (19.8 vs.\ 24.6 tok/s).
CPU package power is nearly identical (90\,W vs.\ 92\,W), so the energy gap is driven almost entirely by latency, yielding $1.22\times$ higher energy per token for the SNN (4,559 vs.\ 3,752\,mJ/tok).
The SNN gains only $1.15\times$ throughput from GPU execution while GPT-2 gains $3.7\times$, showing that the current implementation does not map efficiently to dense GPU kernels.

\paragraph{Neuromorphic projection.}
Only the decay-input projection and FFN down-projection consume spike-valued inputs among the major linear operators; the other major operators consume continuous-valued inputs. All are executed through conventional dense kernels in this implementation.
We report no analytical neuromorphic energy number; \Cref{app:energy_model} states what such an estimate would require.

\subsection{Artifact Scope}
\label{sec:evidence_scope}

\Cref{tab:evidence} summarizes the records supporting each result family.
The public artifacts reproduce checkpoint loading and inference; the training and evaluation results rely on retained experimental records.

\begin{table}[H]
\centering
\caption{Record type supporting each result family.}
\label{tab:evidence}
\small
\begin{adjustbox}{max width=\linewidth}
\begin{tabular}{@{}p{0.30\linewidth}p{0.62\linewidth}@{}}
\toprule
\textbf{Result} & \textbf{Supporting record} \\
\midrule
194M PPL, per-domain results, and ablations & Checkpoints and evaluation outputs for the four SNN runs, the primary dense baselines, and the ablation variants \\
Training dynamics, sparsity, gates, and decay factors & Step-level logs and checkpoint extractions for the four SNN runs \\
Zero-shot, generation, speed, and power & Aggregate evaluation records; per-example zero-shot outputs and complete generated sample sets were not retained \\
0.8B scale-up & Public checkpoint and inference code; internal training and evaluation diagnostic records \\
\bottomrule
\end{tabular}
\end{adjustbox}
\end{table}

LAMBADA uses final-subword scoring (\Cref{sec:zeroshot}), and Distinct-$n$ is reported against GPT-2 201M under one sampling configuration (\Cref{sec:generation}).

\section{Discussion}
\label{sec:discussion}

\subsection{From Pure-Spike Prototypes to Spike-Gated Dual-Path}

Earlier SymbolicLight prototypes~\citep{liu2026symboliclight,liu2026scaling} reported language modeling with high activation sparsity and a quality gap against dense baselines; these observations motivated further study of continuous representations.
SymbolicLight V1 adds windowed local attention on the continuous residual stream, a context MLP head, and a bilingual tokenizer with longer merges.
Among the 0.5B-token configurations, the variant without local attention has the highest PPL, and the static-prior variant also has higher PPL than the full-model reference (\Cref{sec:ablation}). Because the reference uses a different objective and hardware, these observations do not isolate the causal contribution of either component.
The tokenizer merge length is a design choice without an ablation.
The resulting model is a hybrid: softmax attention and a GELU MLP are continuous, while spikes gate the decay-path projection and FFN binarization.

\subsection{Quality Cost in the Primary Comparison}
\label{sec:architecture_tax}

The token-weighted SNN mean is 7.7\% above GPT-2 201M under the shared A800 protocol.
SymbolicLight changes activation type, sequence mixing, recurrent state, parameter allocation, and the number of optimizer updates, so the gap is not a binarization-only tax.
The unweighted domain mean of 29.38 further shows that the 8.904 headline is a Code-heavy average, not a balanced ten-domain quality number.

\subsection{Biological Motivation}

The architecture uses cortical spike-plus-graded-potential organization as a design analogy~\citep{london2005dendritic,gerstner2002spiking}, not as a hardware claim.
Device power readings and derived energy estimates are in \Cref{sec:inference_bench}.

\subsection{Limitations}

Dense GPT-2 baselines are single runs. The 194M evaluation-set separation is unverified, and the component comparisons change objective and hardware. Encoder probes do not measure whole-model sparsity, and post-generation GPU power sampling does not establish integrated energy.
Zero-shot accuracy stays near chance on PIQA and HellaSwag at 200M scale.
Generation metrics measure n-gram diversity, not factual quality.

\section{Conclusion}
\label{sec:conclusion}

We introduced SymbolicLight V1, a spike-gated dual-path language model that reaches token-weighted evaluation PPL 8.88--8.93 (mean 8.904, sample standard deviation 0.019) with mean sampled encoder zero-spike fraction 89.96\%.
Under the shared A800 protocol that mean is 7.7\% above GPT-2 201M; the unweighted mean of the ten domain PPLs is 29.38.
Five zero-shot tasks show no clear accuracy separation at this scale.
A 0.8B run trained for 48.8B tokens yields a loadable checkpoint.
The RTX 2080 Ti throughput is 22.8 versus 91.5 token/s, while post-generation power readings give rough energy estimates of 2,848 versus 905\,mJ/token for GPT-2 201M. Evaluation-set separation remains unverified, so the PPL results cannot establish unseen-data generalization.

\section*{Reproducibility Statement}

The public code repository (\GitHubReleaseURL) provides parameter-compatible model definitions, tokenizer assets, integrity manifests, checkpoint-loading utilities, and minimal generation programs for the 194M and 0.8B models. Checkpoints are distributed through \HFReleaseURL. Together, these artifacts support checkpoint loading, tokenizer compatibility checks, inference-path inspection, and short text generation.

The release does not include the training or evaluation pipelines, optimizer states, data-processing code, or training and validation corpora. It therefore does not enable regeneration of the training, ablation, evaluation, speed, or power results. The architecture and reported hyperparameters are specified in \Cref{sec:arch,tab:config,tab:training}; the public reproducibility claim is limited to checkpoint loading and inference. Verifying generalization additionally requires source identity, a train/evaluation split manifest, and an overlap audit.

\begin{table}[H]
\centering
\caption{SHA-256 manifest for the separately distributed \emph{0.8B} checkpoint and its tokenizer model. These files must not be paired with the 194M 48K tokenizer. Digests are those of the released files.}
\label{tab:sha256}
\scriptsize
\begin{adjustbox}{max width=\linewidth}
\begin{tabular}{@{}lrl@{}}
\toprule
\textbf{Filename} & \textbf{Bytes} & \textbf{SHA-256 prefix} \\
\midrule
\texttt{weights/pytorch/latest-inference.pt} & 3,500,562,525 & \texttt{5edc3b6a6d3fb9cf} \\
\texttt{tokenizer/sl\_tokenizer.model} & 869,691 & \texttt{5b673c90dbcfb8e1} \\
\bottomrule
\end{tabular}
\end{adjustbox}
\end{table}
\noindent Full SHA-256 digests are provided in the release manifests accompanying the code and model-artifact records.

\paragraph{Compute resources.}
\Cref{tab:compute} summarizes the hardware and time budget for the reported 194M-scale training, ablation, and baseline experiments.
Total compute is approximately 430 GPU-hours, with the four primary SNN training runs accounting for the majority.
The 0.8B scale-up run is not included in this compute total.

\begin{table}[H]
\centering
\caption{Compute budget for the reported 194M-scale experiments. Rows marked $^\ast$ use wall-clock times from training logs; the remaining times are estimates. The total excludes the GPT-2 201M run, whose time was not recorded, and the 0.8B scale-up run. The RTX 5090 full-model reference uses a different hardware partition from the A800 experiments.}
\label{tab:compute}
\small
\begin{adjustbox}{max width=\linewidth}
\begin{tabular}{@{}llccc@{}}
\toprule
\textbf{Experiment} & \textbf{Hardware} & \textbf{Runs} & \textbf{Hours/run} & \textbf{GPU-hours} \\
\midrule
SNN training, AuxCE (3B)$^\ast$ & 4$\times$ A800-40GB & 2 & 28.4 & 227 \\
SNN training, noAuxCE (3B)$^\ast$ & 4$\times$ A800-40GB & 2 & 16.1 & 129 \\
GPT-2 124M training (3B)$^\ast$ & 4$\times$ A800-40GB & 1 & 3.5 & 14 \\
GPT-2 201M training (3B) & 4$\times$ A800-40GB & 1 & not recorded & --- \\
Component ablations (0.5B) & 4$\times$ A800-40GB & 3 & ${\sim}$2 & 24 \\
Full model reference (0.5B) & 8$\times$ RTX 5090 & 1 & ${\sim}$1.8 & 14 \\
Surrogate comparison (2K steps) & 4$\times$ A800-40GB & 2 & ${\sim}$0.5 & 4 \\
Supplementary GPT-2 195.9M (3B) & 8$\times$ RTX 5090 & 1 & 2.2 & 18 \\
\midrule
\multicolumn{4}{@{}l}{\textbf{Total (excluding the unrecorded 201M run)}} & ${\sim}$\textbf{430} \\
\bottomrule
\end{tabular}
\end{adjustbox}
\end{table}

\noindent
Inference benchmarks (zero-shot, generation, speed, power) were conducted on a single NVIDIA RTX 2080 Ti GPU and an AMD Ryzen 7 5800X CPU, contributing negligible additional compute.

\bibliographystyle{plainnat}
\bibliography{references_arxiv}

\appendix

\section{Per-Domain Ablation Results}
\label{app:ablation_perdomain}

\Cref{tab:ablation_perdomain} gives per-domain evaluation PPL for the 0.5B-token full model and the three removal variants.

\begin{table}[H]
\centering
\caption{Per-domain evaluation PPL at 0.5B tokens. Full (0.5B) is the matched-budget reference (overall 17.72).}
\label{tab:ablation_perdomain}
\small
\begin{adjustbox}{max width=\linewidth}
\begin{tabular}{@{}lcccc@{}}
\toprule
\textbf{Domain} & \textbf{Full (0.5B)} & \textbf{Static Prior} & \textbf{No Attn} & \textbf{Decay Only} \\
\midrule
Chinese-Reference & 45.29 & 52.42 & 82.92 & 55.15 \\
Chinese-Web       & 385.61 & 415.47 & 592.84 & 430.15 \\
Chinese-General   & 226.42 & 245.98 & 362.38 & 259.06 \\
English-Educational & 21.90 & 26.30 & 45.71 & 25.87 \\
English-Math      & 34.94 & 41.97 & 70.74 & 43.95 \\
English-Reference & 38.13 & 50.96 & 82.99 & 49.88 \\
Math-Web          & 27.22 & 32.64 & 53.72 & 35.46 \\
Code              & 6.00 & 7.21 & 15.04 & 10.14 \\
English-Narrative & 20.22 & 25.53 & 40.55 & 24.46 \\
Chinese-Narrative & 156.44 & 175.16 & 272.87 & 197.40 \\
\midrule
\textbf{Overall} & \textbf{17.72} & \textbf{21.26} & \textbf{38.56} & \textbf{25.27} \\
\bottomrule
\end{tabular}
\end{adjustbox}
\end{table}

\section{Full Gate and Decay Factor Table}
\label{app:gate_full}

\Cref{tab:gate_full} reports the learned gate values $g = \sigma(w_g)$ and mean exponential decay factors $\alpha$ for all 12 layers across all four training runs.

\begin{table}[H]
\centering
\caption{Learned gate values from four runs spanning three seeds and two objectives, with mean decay factors for all 12 layers. Gate $g > 0.5$ favors attention; $g < 0.5$ favors decay.}
\label{tab:gate_full}
\small
\begin{adjustbox}{max width=\linewidth}
\begin{tabular}{@{}ccccccc@{}}
\toprule
\textbf{Block} & \textbf{AuxCE 123} & \textbf{AuxCE 456} & \textbf{noAuxCE 42} & \textbf{noAuxCE 123} & \textbf{Mean $g$} & \textbf{Mean $\alpha$} \\
\midrule
0  & 0.490 & 0.489 & 0.489 & 0.488 & 0.489 & 0.915 \\
1  & 0.526 & 0.520 & 0.521 & 0.535 & 0.526 & 0.929 \\
2  & 0.536 & 0.529 & 0.521 & 0.541 & 0.531 & 0.927 \\
3  & 0.537 & 0.527 & 0.547 & 0.533 & 0.536 & 0.931 \\
4  & 0.556 & 0.568 & 0.556 & 0.541 & 0.555 & 0.933 \\
5  & 0.573 & 0.590 & 0.578 & 0.564 & 0.576 & 0.936 \\
6  & 0.578 & 0.557 & 0.576 & 0.573 & 0.571 & 0.938 \\
7  & 0.575 & 0.610 & 0.606 & 0.580 & 0.593 & 0.941 \\
8  & 0.576 & 0.599 & 0.599 & 0.584 & 0.589 & 0.943 \\
9  & 0.564 & 0.604 & 0.596 & 0.561 & 0.582 & 0.944 \\
10 & 0.632 & 0.576 & 0.574 & 0.624 & 0.602 & 0.948 \\
11 & 0.609 & 0.579 & 0.563 & 0.582 & 0.583 & 0.949 \\
\midrule
\textbf{Mean} & 0.563 & 0.562 & 0.560 & 0.559 & \textbf{0.561} & \textbf{0.936} \\
\bottomrule
\end{tabular}
\end{adjustbox}
\end{table}

The gate progression from ${\sim}0.49$ (Block 0) to ${\sim}0.60$ (Block 10--11) shows a shallow-decay / deep-attention trend in these four runs.
Across-run population standard deviation is below 0.02 for 11 of 12 layers and reaches 0.0264 at Block~10.
The decay factor $\alpha$ increases from 0.915 (Block 0) to 0.949 (Block 11), corresponding to effective memory half-lives of $\log 0.5 / \log \alpha \approx 8$ tokens (shallow) to ${\sim}13$ tokens (deep).

\subsection{Per-Head Decay Heterogeneity}
\label{app:perhead_decay}

\Cref{tab:gate_full} reports the layer-wise mean of the per-head decay factor $\alpha_h = \sigma(\gamma_h)$ averaged over all $H{=}12$ heads.
At a finer granularity, the 12 heads within each layer are not constrained to share a single $\alpha$ value: each head learns its own decay independently, and the table below reports their pooled distribution across training runs.
\Cref{tab:perhead_decay} summarizes the within-layer min/max/range of $\alpha_h$ aggregated over all four training runs (576 measurements total: 4 runs $\times$ 12 layers $\times$ 12 heads).

\begin{table}[H]
\centering
\caption{Per-head decay factor $\alpha_h$ statistics within each layer, aggregated across four training runs (12 heads $\times$ 4 runs $=$ 48 measurements per row). The right column is recomputed from the displayed $\alpha$ extrema as $\log(0.5)/\log(\alpha)$, rounded to one decimal place (tokens).}
\label{tab:perhead_decay}
\small
\begin{adjustbox}{max width=\linewidth}
\begin{tabular}{@{}cccccc@{}}
\toprule
\textbf{Block} & \textbf{Mean $\alpha$} & \textbf{Min $\alpha$} & \textbf{Max $\alpha$} & \textbf{Range} & \textbf{Half-life range (tokens)} \\
\midrule
0  & 0.9149 & 0.9064 & 0.9219 & 0.0155 & 7.1--8.5 \\
1  & 0.9291 & 0.9233 & 0.9347 & 0.0114 & 8.7--10.3 \\
2  & 0.9271 & 0.9183 & 0.9346 & 0.0163 & 8.1--10.2 \\
3  & 0.9306 & 0.9231 & 0.9374 & 0.0143 & 8.7--10.7 \\
4  & 0.9332 & 0.9228 & 0.9423 & 0.0195 & 8.6--11.7 \\
5  & 0.9358 & 0.9274 & 0.9423 & 0.0149 & 9.2--11.7 \\
6  & 0.9381 & 0.9275 & 0.9432 & 0.0157 & 9.2--11.9 \\
7  & 0.9414 & 0.9333 & 0.9549 & 0.0216 & 10.0--15.0 \\
8  & 0.9432 & 0.9323 & 0.9491 & 0.0168 & 9.9--13.3 \\
9  & 0.9437 & 0.9341 & 0.9532 & 0.0191 & 10.2--14.5 \\
10 & 0.9482 & 0.9398 & 0.9559 & 0.0161 & 11.2--15.4 \\
11 & 0.9488 & 0.9402 & 0.9550 & 0.0148 & 11.2--15.1 \\
\bottomrule
\end{tabular}
\end{adjustbox}
\end{table}

The pooled ranges shift from about 7--12 token half-lives in Blocks 0--5 to 9--15 in Blocks 6--11. Their widths fluctuate rather than increasing monotonically. Because the extrema are pooled across four runs, a row does not establish the head-to-head span within any individual trained model. The half-life describes the direct exponential term, not an end-to-end language-memory measurement.

\section{Generation Samples}
\label{app:generation}

\Cref{tab:gen_en,tab:gen_zh} show representative generation samples (sampling decoding, temperature 0.7, top-$k$ 50, 150 tokens) for the EN-Story and ZH-Story domains.
Both examples contain semantic and referential errors. The retained repetition values describe these examples only; lower repetition is not evidence of coherent storytelling.

\begin{table}[H]
\centering
\caption{EN-Story generation (sampling). Prompt: ``\textit{Once upon a time, there was a little girl named Lily who loved to}''}
\label{tab:gen_en}
\small
\begin{adjustbox}{max width=\linewidth}
\begin{tabular}{@{}lp{12cm}c@{}}
\toprule
\textbf{Model} & \textbf{Generated Text (truncated)} & \textbf{Rep.} \\
\midrule
SNN & play in the garden. One day, she found a big cake on it! Lily's favorite dress was so excited to go on the cake. Lily's friend, Lily, said, ``Do you want to go, Lily?'' The cake replied, ``I don't want to go to the cake, but I didn't have anything to do at home.'' Lily learned it all about the cake and was a good kind friend\ldots & 6.1\% \\
\midrule
GPT-2 & play with her friends. One day, Lily's mom came into her room and said, ``I think you are boring!'' Her mom said, ``I want to play with me so much.'' Lily said, ``I don't know how to play with me. I can play with me so much.'' Lily was excited to see what her mom said\ldots & 14.3\% \\
\bottomrule
\end{tabular}
\end{adjustbox}
\end{table}

\begin{table}[H]
\centering
\caption{ZH-Story generation (sampling). Prompt: ``\begin{CJK}{UTF8}{gbsn}从前，有一个小男孩叫小明，他住在一个\end{CJK}''}
\label{tab:gen_zh}
\small
\begin{adjustbox}{max width=\linewidth}
\begin{tabular}{@{}lp{12cm}c@{}}
\toprule
\textbf{Model} & \textbf{Generated Text (truncated)} & \textbf{Rep.} \\
\midrule
SNN & \begin{CJK}{UTF8}{gbsn}村庄，他家里有一只狗，在那里他遇到了一个名叫杰克的朋友们，他找到了一个名叫杰克的家庭，他遇到了一个名叫杰克的弟弟，他们让他有了新的生活。杰克告诉他的好友，他遇到了一个名叫杰克的同事……\end{CJK} & 16.3\% \\
\midrule
GPT-2 & \begin{CJK}{UTF8}{gbsn}神秘的房子，他在那里他迷路了，但他已经找到了这个宝藏。他找到了这个宝藏，他找到了这个宝藏。他遇到了一个神秘的小村庄，他们发现了他的宝藏。他找到了这个宝藏，他找到了宝藏，他找到了宝藏，他找到了宝藏……\end{CJK} & 64.6\% \\
\bottomrule
\end{tabular}
\end{adjustbox}
\end{table}

\section{Requirements for a Neuromorphic Energy Estimate}
\label{app:energy_model}

The decay-input and FFN down-projections consume binary spikes; the FFN up-projection, Q/K/V projections, attention, output projection, and context MLP use continuous values. Zeros alone establish neither operation skipping nor memory-access savings.

We report no analytical neuromorphic advantage. Estimating it requires operator traces, both models' precision and memory-placement assumptions, and validation on a sparse runtime or physical hardware. The RTX 2080 Ti estimates of 2,848 and 905\,mJ/token multiply post-generation power by mean latency. They exclude power integration and whole-system accounting, so they cannot validate such an estimate.

\end{document}